\documentclass[letterpaper, 10 pt, conference]{ieeeconf}

\IEEEoverridecommandlockouts

\usepackage{graphicx}
\usepackage{amsmath}
\usepackage{amsfonts}
\usepackage{hyperref}
\usepackage{siunitx}
\usepackage{float}

\usepackage[ruled, linesnumbered]{algorithm2e}
\usepackage[sorting=none, eprint=false, doi=false]{biblatex}
\renewbibmacro*{url+urldate}{}
\bibliography{citations}

\usepackage{caption}
\usepackage{multirow}

\usepackage[font=small]{caption}
\title{\LARGE \bf HGI-SLAM: Loop Closure With Human and Geometric Importance Features*}

\author{
Shuhul Mujoo$^{1}$ and Jerry Ng$^{2}$%
\thanks{*This work was not supported by any organization}%
\thanks{$^{1}$Shuhul Mujoo is a high school student at Evergreen Valley High School, 3300 Quimby Road San Jose, CA 95148, USA {\tt\small shuhul.mujoo@gmail.com}}%
\thanks{$^{2}$Jerry Ng with the Department of Mechanical Engineering, Massachusetts Institute of Technology, Massachusetts, USA{\tt\small jerryng@mit.edu}}%
}

\begin{document}
\maketitle
\thispagestyle{empty}
\pagestyle{empty}

\begin{abstract}

We present Human and Geometric Importance SLAM (HGI-SLAM), a novel approach to loop closure using salient and geometric features. Loop closure is a key element of SLAM, with many established methods for this problem. However, current methods are narrow, using either geometric or salient based features. We merge their successes into a model that outperforms both types of methods alone. Our method utilizes inexpensive monocular cameras and does not depend on depth sensors nor Lidar. HGI-SLAM utilizes  geometric and salient features, processes them into descriptors, and optimizes them for a bag of words algorithm. By using a concurrent thread and combing our loop closure detection with ORB-SLAM2, our system is a complete SLAM framework. We present extensive evaluations of HGI loop detection and HGI-SLAM on the KITTI and EuRoC datasets. We also provide a qualitative analysis of our features. Our method runs in real time, and is robust to large viewpoint changes while staying accurate in organic environments. HGI-SLAM is an end-to-end SLAM system that only requires monocular vision and is comparable in performance to state-of-the-art SLAM methods.

\end{abstract}

\begin{figure*}[t]
  \centering
  \includegraphics[scale=0.45]{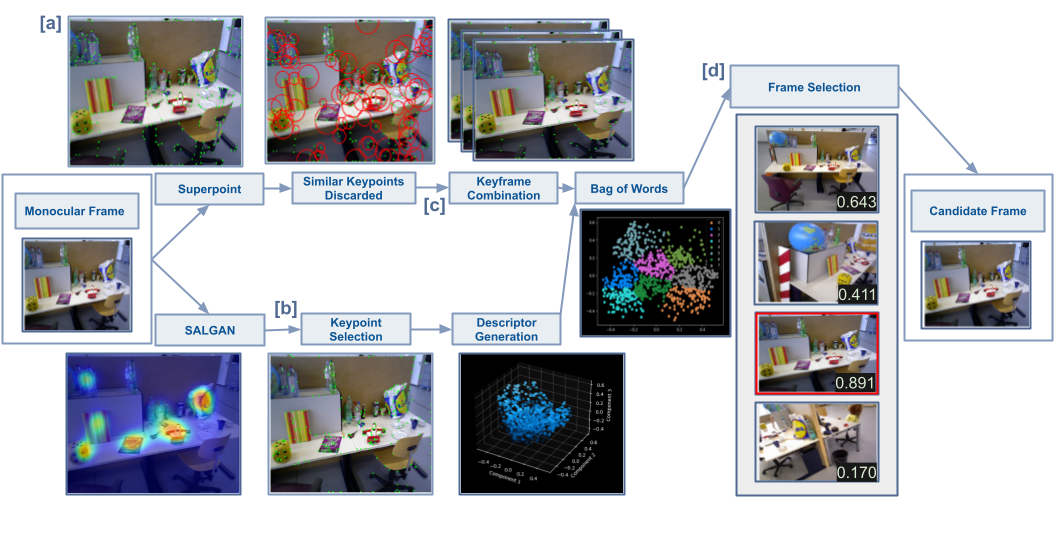}
  \caption{Overview of HGI loop closure detection. First, a monocular frame is passed to the system and fed into two processes, Superpoint (Geometric) [a] and SALGAN (salient). To reduce redundant information, similar geometric keypoints are discarded, and consecutive keyframes are combined [c]. Salient keypoints are generated from the heatmap, and descriptors are generated through a novel modified version of SIFT. Both types of descriptors are processed by a Bag of Words model, which is used to determine the similarity to the target frame. The frame selection [d] chooses the highest similarly score (shown in the bottom right of each image) which becomes the candidate frame. }
  \label{fig:banner}
\end{figure*}

\section{Introduction}

For many years, Simultaneous Localization and Mapping (SLAM) has been the subject of technical research \cite{survey}. But with vast improvements in computer processing speed and the availability of low-cost sensors, SLAM is now used for practical applications in a growing number of fields \cite{survey2}. A basic SLAM system consists of map generation and concurrent robot localization.

One of the most crucial tasks for a SLAM system is to detect when the robot reaches an already seen position in order to correct the map it creates, or loop closure. Loop closure is a key element of SLAM, and without it, local errors quickly build into global errors over time \cite{relocal}. Loop closure is composed of place recognition, determining whether the current frame has already been seen, and loop correction \cite{placerecog}.

Place recognition is characterized as being able to recognize the same place despite significant changes in appearance and viewpoint. Current state-of-the-art SLAM systems, such as ORB-SLAM3 \cite{orbslam3}, rely on using Lidar or depth information in order to perform geometric validation. However, RGBD and stereo cameras are not available in every situation due to their high cost; they are several times more expensive than their monocular counterparts \cite{gated}. The ability to only rely on monocular input makes SLAM systems even more versatile. For example, in rural areas, small monocular cameras are ideal due to their low power consumption and availability.

There have been many proposed methods for feature detection, but these have been split mainly into two categories. Methods like SuperPoint \cite{superpoint}, SIFT \cite{sift}, and ORB \cite{orb} are good at detecting geometric features, but fail in organic environments \cite{survey3}. On the other hand, methods like SalientDSO \cite{sdso}, Salient Point Detection \cite{spd}, and Feature Based SLAM \cite{fbs}, determine points of human interest, but only target key objects. In order to create loop closures that are reliable after large viewpoint shifts and in object-less environments, a combination of these two types of features can be used for loop closure. Some multiple feature detectors already exist \cite{fusion}, but these methods rely on two features from the same category. By creating features that represent both human and geometric importance (HGI), the benefits of both can be realized. 

These ideas motivate our method, HGI-SLAM. Newer feature detection methods use machine learning based methods \cite{liftslam}, and we continue with this technique. Although these methods can be slower than algorithmic methods, they outperform them in terms of accuracy and viewpoint invariance. Despite this, our method can be run in real time due the descriptor generation occuring in a separate thread.

We present HGI-SLAM, a novel combination of features for more accurate loop closures. Using only monocular camera data, we have made it accessible as possible. The key contributions of this paper are:

\begin{itemize}
\item A novel combination of geometric and salient feature generation for place recognition
\item A modified descriptor generation algorithm based on SIFT
\item An end-to-end framework using loop closure detections in a multi-threading setup. We pass them to the ORB-SLAM2 backbone by overriding multiple keyframes
\end{itemize}

In addition to these, we also provide experimental results on the KITTI \cite{kitti} and EuRoC \cite{euroc} datasets to demonstrate the improved performance of the proposed loop closure approach.

The rest of the paper is organized as follows: Sec. II presents related work to our own and the core elements that we built on. Sec. III describes our method of feature extraction, starting from raw images to loop closure detection. This section details the majority of our novel contribution. Qualitative results of our novel loop closure detection and combined system are given in Sec. IV. We also provide a qualitative analysis of features and timing results in Sec. IV. We conclude the paper in Sec. V with a summary of our proposed framework and future scope and applications.

\section{Related Work}

Many past methodologies for loop closure rely on specific sets of features along with a bag of words model (BoW). Exceptions to BoW exist \cite{rsom}, but bag of words has been shown to be useful for retrieving similar images for loop closure detection \cite{survey3}. 

Feature detection methods such as SIFT \cite{sift}, SURF \cite{surf}, and FAST \cite{fast} detect sparse keypoints, but other types of feature detectors have been proposed (see \cite{line} for line based features). These features have been integrated into slam systems. For example, ORB-SLAM2 \cite{orbslam2} uses FAST for keypoint selection and LDSO \cite{ldso} uses DSO. Due to the speed at which they can be detected and matched, our method uses sparse keypoints as features.

More recent methods of loop closure include deep-learning methods, such as SymbioLCD \cite{symbioLCD} or Online VPR \cite{onlineVPR}. These methods tend to more computationally expensive, but can perform better \cite{survey4}. There are also methods for Lidar based slam such as OverlapNet \cite{overlapnet} and LCDNet \cite{lcdnet}. These methods require Lidar while our method focuses on using only a monocular camera.

HGI-SLAM is built on two major systems. The first is Superpoint \cite{superpoint}. Superpoint is a self supervised framework for interest point and descriptor detection. It is a fully convolutional model that is largely viewpoint invariant. The interest point detector, MagicPoint, preforms a lot better than traditional corner detection approaches like FAST \cite{fast} or Harris \cite{harris} \cite{magicpoint}. SuperPoint tends to produce more dense and correct matches compared to LIFT \cite{lift}, SIFT \cite{sift} and ORB \cite{orb}.

The second system is SalientDSO, \cite{sdso}. SalientDSO (SDSO) is a way to incorporate semantic information in the form of visual saliency into Direct Sparse Odometry. SDSO generates a saliency heatmap using SALGAN \cite{salgan}, which introduced the use of Generative Adversarial Network (GAN) for saliency prediction. The heat map is then filtered based on scene parsing, and interest points are extracted from the image. SDSO is robust even with a small number of features, producing low drift in organic environments.

Both of these systems work exceedingly well in their optimal environment. HGI-SLAM combines the best of both systems while only requiring monocular vision.

\begin{figure}[h]
  \centering
  \includegraphics[scale=0.38]{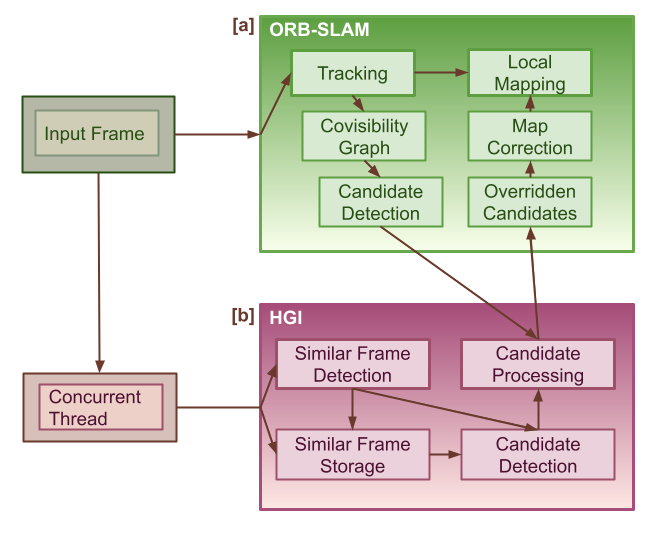}
  \caption{HGI-SLAM, a combination of HGI features and ORB-SLAM2. The input frame from a sequence or camera is passed to ORB-SLAM2 [a] and a concurrent thread running HGI [b] (shown in more detail in Fig. \ref{fig:banner}). When there is a loop closure, the candidate frames produced by HGI override the frames in ORB-SLAM2. These frames are then used for map correction and tracking.}
  \label{fig:system}
\end{figure}

\section{Method}

The framework for HGI-SLAM contains the following steps: First, geometric and salient features are extracted in the form of keypoints and descriptors. This involves running Superpoint \cite{superpoint} and SDSO \cite{sdso}, then processing and optimizing these features. Since SDSO does not output descriptors, these were computed with a SIFT \cite{sift} like algorithm on the interest points and the original heatmap from SALGAN \cite{salgan}. Next, the features are combined to train a BoW model to generate a vocabulary for future reference. Finally, the loop closures are detected in a concurrent thread to ORB-SLAM2 and keyframes in ORB-SLAM2 are overridden.

\subsection{Geometric features}

We adopt the keypoints and descriptors from Superpoint by running them on the current frame. Superpoint \cite{superpoint} uses a fully convolutional neural network and homographic adaption to find keypoints and descriptors. These points target corners and geometric features in the input image. They are also scale and rotation invariant to the extent found in the training dataset.

However, when Superpoint is run on an image with high texture or lots of contrast, the keypoints cluster around that part of the image. In order to diversify the keypoints, we remove keypoints that are nearby and have similar descriptors.

Let the set of keypoints, $\mathcal{K} =
\{\mathbf{k}_1, \dots, \mathbf{k}_N\}$, where $\mathbf{k}_i \in \mathbb{R}^2 $. First, we compute the nearest keypoints as follows:

\begin{equation}
\mathcal{M}_i = \{\mathbf{k} \in \mathcal{K} \mid {\lVert \mathbf{k}_i-\mathbf{k} \rVert}^2 < T \} \label{eq:nearkeypoints}
\end{equation}
where $T$ is the threshold distance. If $|\mathcal{M}_i| \leq N$ where $N$ is some maximum amount of nearby keypoints then $\mathbf{k}_i$ is kept. Otherwise the average cosine similarity \cite{cossim} $\bar s_i$ between the corresponding descriptors $\mathcal{D}_i$ of keypoints in $\mathcal{M}_i$ is computed:

\begin{equation}
\bar s_i = \frac{1}{|\mathcal{D}_i|}\sum_{\mathbf{d} \in \mathcal{D}_i} \frac{\mathbf{d} \cdot \mathbf{k}_i}{||\mathbf{d}||||\mathbf{k}_i||}
\label{eq:similarity}
\end{equation}
The keypoints that have an average similarity greater than a minimum $s_{\mathrm{min}}$ are discarded. This process removes keypoints that are both nearby and similar. The values that were used were $T=50$, $N=10$, and $s_{\mathrm{min}}=0.6$.

The descriptors of the keypoints go through one last optimization step before being used in the BoW model, unlike prior art. Three consecutive frames' keypoints are combined together (overlayed) and used for only the center frame. This saves storage space and removes unnecessary information, as consecutive frames are usually interchangeable. However, in order to retain information from these frames, similarity is again calculated using Eq. \ref{eq:similarity}, this time between descriptors from the separate frames. Similar keypoints are removed until the number of keypoints equals the number of keypoints from the original output from Superpoint.

\subsection{Salient features}

The next step is to use the saliency heatmap generated by SALGAN \cite{salgan} to create keypoints and descriptors. In the original paper, SalientDSO (SDSO) \cite{sdso} uses this heatmap to create keypoints but not descriptors. The heatmap is used with PSPnet to determine the semantic label of objects in the scene. Then keypoints are selected by splitting this image into patches, and pixels with the highest gradient are selected. In our method, we adopt a similar method to that of SDSO.

However, instead of using PSPNet \cite{psp}, we run a modified version of the point selection algorithm. This is because the limited number of semantic labels of PSPNet limits SDSO to indoor use.

To extract keypoints from the saliency heatmap, first we compute the gradient of the entire heatmap and store it into two matrices $\nabla \mathbf{G}_m$ and $\nabla \mathbf{G}_o$, the magnitude and orientation of the gradients respectively. Then a random $8 \times 8$ patch $M_i$ of the original image $I$, is selected. The sampling weight is almost the same as SDSO except we replace the median with the average of the gradient in that patch. We do this in order to keep the sampling probability consistent and to avoid the use of a region-adaptive threshold.

\begin{equation}
w_i = \frac{1}{64}\sum_{j \in M_i} \nabla\mathbf{G}_m [j]
\label{eq:samplingweight}
\end{equation}
Using Eq. \ref{eq:samplingweight} we compute the probability of a patch $M_i$ being sampled as:
\begin{equation}
{P}(M_i) = \frac{w_i}{\sum_{m \in \mathbf{I}}w_m}
\label{eq:probpatch}
\end{equation}


\begin{algorithm} [b]
\caption{Salient/SIFT descriptor generation}\label{alg:cap}
\label{alg:saldesc}
\DontPrintSemicolon
\KwIn{Input image $I$, keypoints $\mathbf{k}$}
\KwOut{Descriptors for each keypoint $\mathbf{d}$ }
Blur $I$ with Gaussian kernel of size $5$ \\
$\mathbf{d} \gets \{\emptyset\}$ \\
$i \gets 0$ \\
\While{$i$  $< \mathrm{length}(\mathbf{k})$}{
Initialize $v$ of length $128$ \\
$k_i \gets \mathbf{k}[i]$ \\
Select $16 \times 16$ region $M$ centered at $k_i$ \\
Compute $\nabla \mathbf{M}_\mathrm{magnitude}$ and $\nabla \mathbf{M}_\mathrm{orientation}$ \\
Split $M$ into $4 \times 4$ blocks \\
\For{$\text{each block } R \text{ in } M$}{
Compute histogram $h$ of orientations \\
Apply shifted cubic interpolation to $h$ \\
Append $h$ to $v$ \\
}
$\mathbf{d}[i] \gets v$
}
\end{algorithm}

After a patch has been selected, we subdivide three times and select points with higher gradient thresholds at every level. This is identical to the SDSO method. \cite{sdso}

Now that we have a set of keypoints $\mathbf{k}$ for each image, we are left with creating the descriptors. To do this, we adapt part of the SIFT \cite{sift} algorithm into a novel descriptor generator. The original image is blurred with a Gaussian kernel of size five. Then for each keypoint a $16 \times 16$ by region $M$ is selected of the gradient magnitude and orientation producing $\nabla \mathbf{M}_m$ and $\nabla \mathbf{M}_o$. For each of the sixteen $4 \times 4$ regions $R$ in $M$ a histogram $H$ is generated. The bins are the orientations split into eight different directions, and the values are the magnitudes of the gradients.

\begin{equation}
b = \left \lfloor{ \frac{R_{o_i}}{\frac{\pi}{4}} }\right \rfloor, R_{m_i} = H[b]
\label{eq:bins}
\end{equation}

The histogram is then smoothed using shifted cubic interpolation. Let $c(x)$ be the cubic interpolation of the histogram. Then to smooth the histogram using a weight parameter, $w = 0.3$,
\begin{equation}
H[i] = \frac{c(i-w) + c(i+w)}{2}
\label{eq:shiftedcubic}
\end{equation}

Finally, each of the 128 raw values generated from the descriptor values for the keypoint are then normalized linearly to $[0,1)$, multiplied by 255, and rounded to integers. This means that the descriptors are a 128 length array where  for each $v_i \in \mathbf{v}$, $v_i \in \{0,1,\dots, 255\}$.

These descriptors do not go through the same optimization step as the superpoint descriptors. Instead, these descriptors are only computed every three frames and then passed to the BoW model. A summary of salient descriptor generation is given in Algorithm \ref{alg:saldesc}. 


\subsection{Loop Closure with Bag of Words}

The set of salient and geometric descriptors is used with a BoW model to detect loop closures. The two types of features go into separate BoW models, due to their different representations and lengths. The BoW model provides an efficient lookup of the closest frames to the current frame as well as their distances. Let $d_s$ and $d_g$ correspond to the distance of the closest frames from the salient and geometric models, respectively. We compute the similarity $s$ between the candidate frame and the current frame $c$ as follows:
\begin{equation}
F(x) = 1 - \exp(-x^{-1})
\label{eq:helpfunc}
\end{equation}
\begin{equation}
s = F(|d_s-d_g|) \cdot F(d_s w_s + d_g w_g)
\label{eq:framesim}
\end{equation}
where $w_s = 0.3$ and $w_g = 0.7$ are weights that determine the relative contribution of salient and geometric features to the loop closure. If $|d_s-d_g| \gg 0$ then the similarly is lowered to account for the difference, i.e. saliency and geometric features that agree are better. Frames with a similarity above a similarity threshold $s_\mathrm{th} = 0.82$ are marked as loop closures.

The last step is to skip loop closure detections of nearby frames ($\approx10$ frames apart), and frames that have already been closed before.

\begin{figure*}[ht]
  \centering
  \includegraphics[scale=0.36]{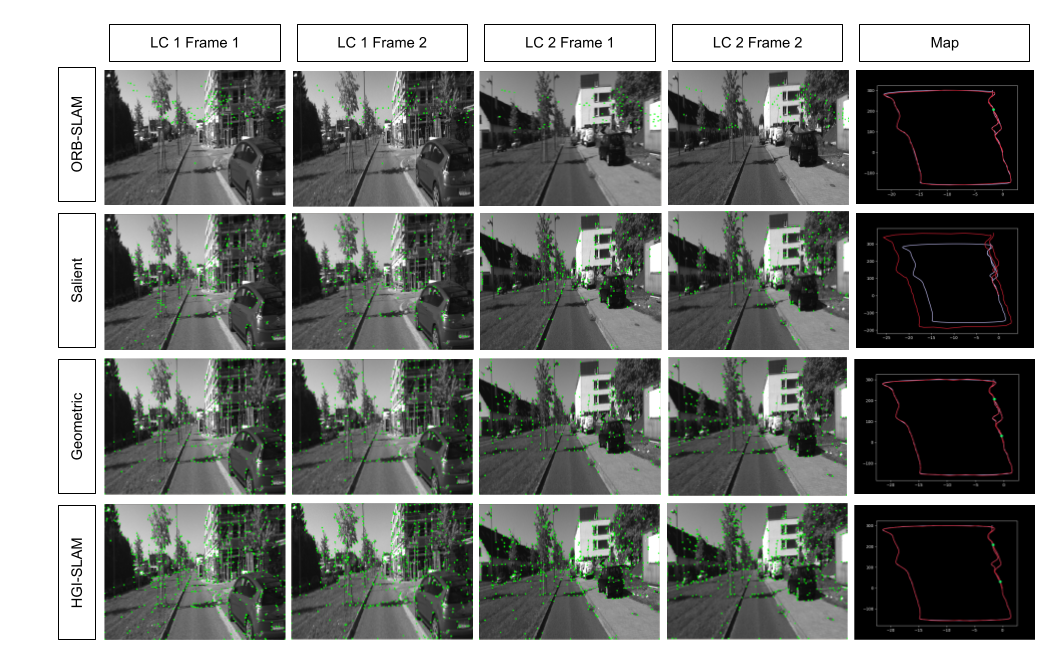}
  \caption{Loop closures detections on sequence 06 of KITTI. There are two viable loop closures and two pairs of frames shown (e.g. LC 1 Frame 1 is the first frame exposed of the the first loop closure). The map generated by each method is at the right. The ground truth is shown in white, the path generated in red, and the loop closure locations in green. ORB-SLAM2 was only able to detect one loop closure, both geometric and HGI were able to detect both, and HGI had the lowest ATE. Note that salient features alone were not able to detect either loop closure due to the instability of the points chosen. }
  \label{fig:maps}
\end{figure*}

\subsection{End-to-end framework}

In order to create a complete SLAM system, we combine our loop closure with ORB-SLAM2. We start a concurrent thread loop closing thread, and modify the default thread. The current frame is passed to HGI, which detects loop closures and delivers the candidate frames back to ORB-SLAM2. This process is shown in Fig. \ref{fig:system}.

After a loop closure is detected, the three ORB-SLAM2 keyframes before the target frame are replaced with frames processed by HGI. To store HGI frames efficiently, only frames that have similarity (computed using Eq. \ref{eq:framesim}) less than $s_\mathrm{th}/2$ when compared to the last framed added, are stored. Loop closures can be attemped even if HGI frames are not stored, relying on ORB features as a fallback.

The loop closure of HGI combined with ORB-SLAM2 is HGI-SLAM, a complete SLAM framework with accurate loop closure detection.

\section{Results}

Our method was evaluated with two metrics. First, we present a quantitative analysis of HGI loop detection in terms of precision and recall. Then we compare the entire HGI-SLAM to its constituent parts to show the improvement of combining feature types. For both parts we use the KITTI \cite{kitti} sequences with loop closures, and all EuRoC \cite{euroc} sequences. The next section provides a qualitative analysis, along with a direct similarity \cite{cossim} comparison of feature types. Finally, we compare the runtime of our method against the base ORB-SLAM2 system.

\subsection{Evaluation of Loop Closure Detection}

For the following two evaluations we use the KITTI \cite{kitti} and EuRoC \cite{euroc} datasets. The KITTI dataset contains stereo sequences recorded from a car in urban and highway environments. Of the stereo images, we use only the first image to provide a monocular input. We also only ran the evaluation on sequences 00, 02, 05, 06, 07, and 09 as they contain loops. The EuRoC dataset contains 11 stereo sequences of different rooms and a large industrial environment. We again only use the first image sequences, and skip sequence V2-03 due to severe motion blur. We ran each method on these sequences to predict loop closure frames, and recorded the average precision and recall for each.

Precision and recall are defined as:
\begin{equation}
\mathrm{Precision} = \frac{TP}{TP + FP}, \text{ }\mathrm{Recall} = \frac{TP}{TP + FN}
\label{eq:prerec}
\end{equation}
where TP, FP, FN are true positives, false positives, and false negatives respectively.

We compared HGI to other monocular methods for loop closure detection, SymbioLCD \cite{symbioLCD} and Online VPR \cite{onlineVPR}. ORB-DBOW2 is also shown as a baseline. Table \ref{tab:prerecKitti} and Table \ref{tab:prerecEuroc} show that HGI had the highest average recall and precision for both datasets. Further, HGI has superior recall on most sequences, with SymbioLCD preforming slightly better on 07 of KITTI and V1-01 of EuRoC. Sequence V1-02 was the only sequence where HGI had both lower recall and precision, which is discussed in Section IV-B. Some specific frames and their keypoints are shown in Fig. \ref{fig:maps}.

\begin{table}[ht!]
\caption{Precision and Recall Comparison on KITTI (\si{\%})}
\label{tab:prerecKitti} \begin{center} \begin{tabular}{c c c c c c c c c}
\hline \hline 
&\multicolumn{2}{c}{ORB-DBoW2} & \multicolumn{2}{c}{SymbioLCD} & \multicolumn{2}{c}{Online VPR} & \multicolumn{2}{c}{HGI} \\
Sequence & Pre. & Rec. & Pre. & Rec. & Pre. & Rec. & Pre. & Rec. \\
\hline 
00 & 57.70 & 51.38 & \textbf{93.49} & 84.72 & 75.07 & 74.51 & 86.50 & \textbf{87.96}\\
02 & 57.02 & 54.90 & 86.69 & 90.32 & 73.64 & 88.89 & \textbf{92.75} & \textbf{92.96}\\
05 & 67.59 & 54.39 & \textbf{87.34} & 83.59 & 77.07 & 71.48 & 85.86 & \textbf{91.77}\\
06 & 68.55 & 60.11 & 87.07 & 77.92 & 80.77 & 76.01 & \textbf{94.41} & \textbf{92.40}\\
07 & 52.12 & 55.96 & 83.19 & \textbf{92.89}  & 84.71 & 73.15 & \textbf{95.81} & 92.45\\
09 & 64.13 & 57.28 & 90.40 & 86.35 & 83.32 & 79.54 & \textbf{91.03} & \textbf{94.72}\\
\hline
Average & 61.15 & 55.67 & 88.03 & 85.97 & 79.10 & 77.26 & \textbf{89.39} & \textbf{91.85} \\
\hline \hline
\end{tabular} \end{center} \end{table}

\begin{table}[ht!]
\caption{Precision and Recall Comparison on EuRoC (\si{\%})}
\label{tab:prerecEuroc} \begin{center} \begin{tabular}{c c c c c c c c c}
\hline \hline 
&\multicolumn{2}{c}{ORB-DBoW2} & \multicolumn{2}{c}{SymbioLCD} & \multicolumn{2}{c}{Online VPR} & \multicolumn{2}{c}{HGI} \\
Sequence & Pre. & Rec. & Pre. & Rec. & Pre. & Rec. & Pre. & Rec. \\
\hline 
MH-01 & 60.87 & 63.36 & 86.56 & 85.35 & 87.08 & 72.72 & \textbf{97.26} & \textbf{88.01}\\
MH-02 & 60.41 & 60.31 & 84.53 & 85.34 & 77.83 & 80.31 & \textbf{85.62} & \textbf{92.04}\\
MH-03 & 59.67 & 64.06 & 87.28 & 94.01 & 78.71 & 74.32 & \textbf{88.04} & \textbf{96.89}\\
MH-04 & 62.20 & 65.17 & \textbf{86.10} & 81.31 & 76.15 & 73.74 & 82.75 & \textbf{94.09}\\
MH-05 & 59.45 & 62.26 & 88.14 & 81.91 & 78.14 & 73.85 & \textbf{99.15} & \textbf{96.41}\\
V1-01 & 64.51 & 61.76 & 90.28 & \textbf{85.51} & 78.69 & 79.64 & \textbf{93.41} & 85.31\\
V1-02 & 50.48 & 56.02 & \textbf{94.73} & \textbf{93.82} & 82.06 & 81.63 & 87.51 & 90.22\\
V1-03 & 52.17 & 63.80 & 91.02 & 81.02 & 72.60 & 71.93 & \textbf{96.65} & \textbf{98.76}\\
V2-01 & 56.02 & 63.62 & 86.44 & 90.87 & 75.85 & 80.32 & \textbf{89.76} & \textbf{96.09}\\
V2-02 & 54.95 & 56.10 & 89.73 & 89.82 & 74.85 & 76.90 & \textbf{98.28} & \textbf{93.84}\\
\hline
Average & 58.07 & 61.65 & 88.47 & 86.89 & 77.20 & 76.54 & \textbf{91.84} & \textbf{92.16} \\
\hline \hline
\end{tabular} \end{center} \end{table}

\subsection{Evaluation of Complete System}

To evaluate the entire HGI-SLAM system, we ran each method on the same sequences as in Section IV-A, and recorded the absolute trajectory error.

Absolute trajectory error is defined as  

\begin{equation}
ATE_\mathrm{rmse} = \left(\frac{1}{n} \sum_{i=1}^n || \mathbf{p}_i - \mathbf{g}_i ||^2 \right)^\frac{1}{2}
\label{eq:rmse}
\end{equation}
where $\mathbf{p}_i$ is the predicted position and $\mathbf{g}_i$ is the ground truth. \cite{robust}

We compared HGI-SLAM to its constituent parts (salient, geometric) by using only that feature type for loop closure detection. ORB-SLAM2 on monocular input is shown as a baseline. Table \ref{tab:kittiATE} and Table \ref{tab:eurocATE} show that HGI-SLAM has comparable ATE to ORB-SLAM2 on most sequences. HGI-SLAM outperforms ORB-SLAM2 in sequences 00, 02, 06, and 09. In sequence 05 and 07, both methods detect nearly all the loop closures, resulting in close ATE. Scale drift occurs in all of the methods due to the monocular input, but the loop corrections in HGI-SLAM improve the overall performance. 

Neither salient nor geometric features individually preformed better than HGI. Using only salient features tends to produce large tracking errors due to the limited salient objects in some scenes. Note that in sequence 09 of KITTI and V1-02 of EuRoC, salient features alone could not complete the tracking. This does not affect the performance of HGI-SLAM because salient features serve as a supplement to geometric features, which work well in most environments. Geometric features largely had a ATE slightly higher than HGI-SLAM or ORB-SLAM2.

\begin{table}
\caption{ATE on KITTI sequences with loop closures (\si{\meter})}
\label{tab:kittiATE} \begin{center} \begin{tabular}{c c c c c}
\hline \hline
Sequence & ORB-SLAM2 & Salient & Geometric & HGI-SLAM\\
\hline
00 & 10.37 & 127.6 & 10.23 & \textbf{7.65}\\
02 & 23.30 & 89.62 & 19.64 & \textbf{13.67}\\
05 & \textbf{5.38} & 91.31 & 49.64 & 8.31\\
06 & 20.93 & 58.97 & 15.92 & \textbf{11.85}\\
07 & \textbf{9.70} & 67.56 & 10.23 & 10.48\\
09 & 38.29 & - & 21.60 & \textbf{20.57}\\
\hline \hline
\end{tabular} \end{center} \end{table}

\begin{table}
\caption{ATE on EuRoC sequences (\si{\meter})}
\label{tab:eurocATE} \begin{center} \begin{tabular}{c c c c c}
\hline \hline
Sequence & ORB-SLAM2 & Salient & Geometric & HGI-SLAM\\
\hline 
MH-01 & 0.149 & 0.343 & 0.187 & \textbf{0.123}\\
MH-02 & 0.130 & 0.449 & 0.204 & \textbf{0.102}\\
MH-03 & 0.142 & 0.878 & \textbf{0.082} & 0.098\\
MH-04 & 0.181 & 0.741 & 0.211 & \textbf{0.157}\\
MH-05 & \textbf{0.119} & 0.672 & 0.154 & 0.142\\
V1-01 & 0.074 & 1.038 & 0.142 & \textbf{0.066}\\
V1-02 & \textbf{0.046} & - & 0.091 & 0.124\\
V1-03 & 0.096 & 0.199 & 0.108 & \textbf{0.065}\\
V2-01 & \textbf{0.055} & 0.507 & 0.129 & 0.125\\
V2-02 & \textbf{0.156} & 0.269 & 0.164 & 0.161\\
\hline \hline
\end{tabular} \end{center} \end{table}

\subsection{Qualitative Analysis of Features}

\begin{figure}
  \centering
  \includegraphics[scale=0.25]{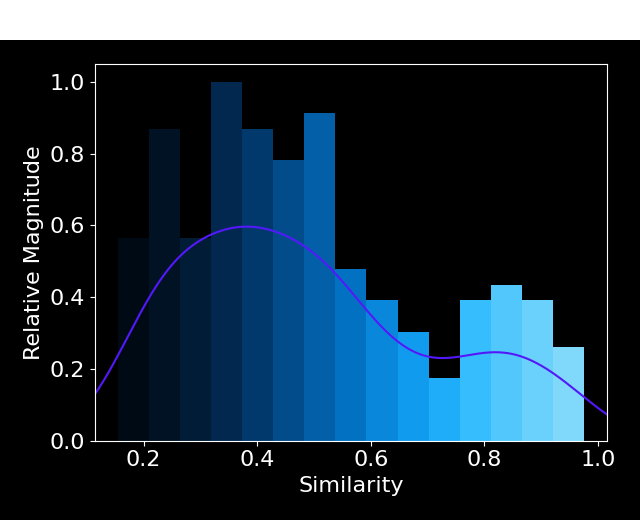}
  \caption{Similarity between geometric and salient features on a sample sequence (KITTI 02).}
  \label{fig:similarity}
\end{figure}

The claim in this paper is that the combination of geometric and salient features results in more accurate loop closure detection. The intuition behind this claim is that saliency represents information about objects in the scene, and geometric features represent high-level position information. This combination results in detections that can rely on two different types of features, providing more robustness and versatility. Fig. \ref{fig:similarity}, captures the uniqueness of each feature type, as they are mostly dissimilar by cosine similarity \cite{cossim}. 

Both salient features and geometric features were collected, and a histogram was created based on the similarity score between them. The graph is scaled so that largest number of elements has a magnitude of one. Fig \ref{fig:similarity} has two peaks of relative magnitude of similarity.  The larger one centered near the left is due to the difference in the type of features. The smaller peak to the right is caused by the alignment of features in certain situations, specifically when salient objects are in front of a mostly featureless background. 

We show through quantitative and qualitative analysis that HGI-SLAM detects loop closures better than by using either feature alone, and better than bare ORB-SLAM2. Our proposed method can detect loop closures in situations with more organic features, which is due to the saliency component. 

\subsection{Runtime Analysis}

In order to complete the evaluation of HGI-SLAM, we present timing results in Table \ref{tab:timings} based on sequences mentioned in the previous section.

The timings in Table \ref{tab:timings} represent the average runtime of each thread. The original tracking thread was only modified to collect data on the system. The original loop thread is where the loop closure frames are replaced, as described in Section III-D. The HGI-loop detection thread is where the loop detection of our method occurred.

Overall, our method does not significantly slower the ORB-SLAM2 base, and is capable of detecting loop closures in real-time. The slowest parts of HGI are the heatmap generation from SALGAN, and descriptor generation, described in Section III-B. The complete system HGI-SLAM runs at 30fps on an Intel® Core i7-10510U CPU @ 1.80GHz with a NVIDIA® GeForce® MX230 graphics card.

\begin{table}[ht!]
\caption{Timings results of each thread, average (\si{ms})}
\label{tab:timings} \begin{center} \begin{tabular}{c|c c c c}
\hline \hline
Thread & ORB-SLAM2 & HGI-SLAM\\
\hline
ORB-SLAM2 Tracking (Barely Modified) & 42.77 & 48.15\\
ORB-SLAM2 Loop (Heavily Modified) & 96.34 & 104.90\\
HGI Loop Detection & - & 86.29\\
\hline \hline
\end{tabular} \end{center} \end{table}

\section{Conclusion}

We have introduced an approach to loop closure using human salient and geometric features, HGI-SLAM. By combining geometric and salient features, our method is able to accurately detect loop closures using either objects or contours. In order to do this, we created a novel descriptor generation method and a fully integrated SLAM system based on ORB-SLAM2 \cite{orbslam2}. By postprocessing loop detections, we removed many false positives and optimize the storage of keyframes. We provide quantitative evaluations of HGI on the KITTI \cite{kitti} and EuRoC \cite{euroc} datasets to show the benefits of this approach. Furthermore, HGI-SLAM is better than either using either type of feature alone, as shown in the evaluation of the complete system. We corroborate with a qualitative analysis of the salient and geometric features.

Lastly, we believe that HGI-SLAM can be extended to further improve its loop closure. Combining the information of neighboring keyframes after loop closure may increase robustness to motion blur. Training the model in a larger variety of environments could further extend its versatility.

\printbibliography

\end{document}